%% file: main.tex
\documentclass[runningheads]{llncs}

\usepackage{eccv}

\usepackage{eccvabbrv}

\usepackage{graphicx}
\usepackage{booktabs}
\usepackage{utfsym} % for check and cross mark, zhuhao

\usepackage[accsupp]{axessibility}  % Improves PDF readability for those with disabilities.

\usepackage{hyperref}

\hypersetup{
    colorlinks,
    citecolor=eccvblue,
    filecolor=black,
    linkcolor=black,
    urlcolor=eccvblue
}

\usepackage{orcidlink}

\usepackage{marvosym} %add by zhuhao, for envelope symbo

\newcommand{\datasetname}{{\sc EmoTalk3D }}

\begin{document}

\title{EmoTalk3D: High-Fidelity Free-View Synthesis \\ of Emotional 3D Talking Head} 

\titlerunning{EmoTalk3D}

\author{Qianyun He$^{1}$, Xinya Ji$^{1}$, Yicheng Gong$^{1}$, Yuanxun Lu$^{1}$, Zhengyu Diao$^{1}$, Linjia Huang$^{1}$, Yao Yao$^{1}$, Siyu Zhu$^{2}$, Zhan Ma$^{1}$, Songcen Xu$^{3}$, Xiaofei Wu$^{3}$, \\ Zixiao Zhang$^{3}$, Xun Cao$^{1}$, Hao Zhu$^{1}$$\textsuperscript{\Letter}$}

\authorrunning{Q.~He et al.}

\institute{$^1$ State Key Laboratory for Novel Software Technology, Nanjing University, China\\
$^2$ Fudan University, Shanghai, China \quad
$^3$ Huawei Noah's Ark Lab
}

\maketitle

\begin{figure*}[t]
  \centering
  \includegraphics[width=1.0\linewidth]{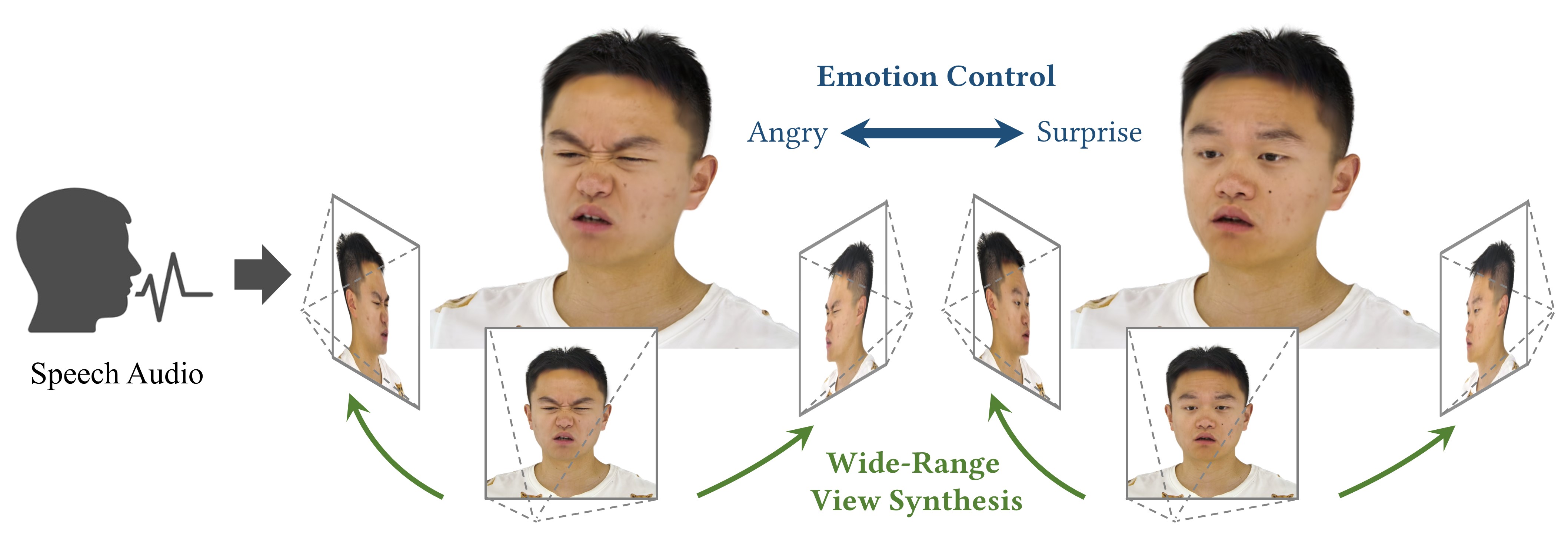}
  \setlength{\abovecaptionskip}{0.1cm}
  \setlength{\belowcaptionskip}{-0.4cm}
  \caption{Given a speech signal, our method can synthesize \textit{high-fidelity}, \textit{emotion-controllable} talking head that can be rendered over a \textit{wide range} of viewing angles.
  }
  \label{fig:teaser}
\end{figure*}

\input{secs/0_abs}
\input{secs/1_intro}

\input{secs/2_related}
\input{secs/3_dataset}
\input{secs/4_method}
\input{secs/5_exp}
\input{secs/6_con}

% \clearpage
\bibliographystyle{splncs04}
\bibliography{mybib}
\end{document}

%% file: secs/0_abs.tex
\begin{abstract}
  We present a novel approach for synthesizing 3D talking heads with controllable emotion, featuring enhanced lip synchronization and rendering quality.  Despite significant progress in the field, prior methods still suffer from multi-view consistency and a lack of emotional expressiveness. To address these issues, we collect \datasetname dataset with calibrated multi-view videos, emotional annotations, and per-frame 3D geometry.  By training on the \datasetname dataset, we propose a \textit{`Speech-to-Geometry-to-Appearance'} mapping framework that first predicts faithful 3D geometry sequence from the audio features, then the appearance of a 3D talking head represented by 4D Gaussians is synthesized from the predicted geometry. The appearance is further disentangled into canonical and dynamic Gaussians, learned from multi-view videos, and fused to render free-view talking head animation.  Moreover, our model enables controllable emotion in the generated talking heads and can be rendered in wide-range views.  Our method exhibits improved rendering quality and stability in lip motion generation while capturing dynamic facial details such as wrinkles and subtle expressions.  Experiments demonstrate the effectiveness of our approach in generating high-fidelity and emotion-controllable 3D talking heads. The code and \datasetname dataset are released in \url{https://nju-3dv.github.io/projects/EmoTalk3D}.

  \keywords{Talking head, audio-driven generation, emotion synthesis, free-view synthesis, 3D Gaussian splatting}
\end{abstract}

%% file: secs/1_intro.tex
\section{Introduction}
\label{sec:intro}

3D Talking heads refer to synthesizing a person-speaking animation given a speech, which has high applicability in various scenarios, such as digital humans, chatting robots, and virtual conferences.  
The core challenge of this task lies in accurately mapping speech signals to 3d lip movements, facial expressions, and, potentially, emotions.  So this task is highly complex and demands meticulous techniques along with sophisticated algorithms.

Despite numerous studies dedicated to the study of 3D talking heads, state-of-the-art methods still encounter evident challenges. In terms of rendering quality, there remains ample scope for enhancing lip synchronization accuracy. Additionally, the intricate facial expressions, including wrinkles and subtle expressions, are not accurately synthesized. These deficiencies result in artifacts within the rendered animations, thereby diminishing their overall fidelity. Furthermore, animations generated by previous 3D talking head methods often overlook emotional expression, thus restricting users' capacity for emotional communication. A key impediment to the research of emotional 3D talking heads is the lack of speech and 4D head datasets that incorporate emotion annotations. Though multi-view video datasets for emotion annotation are available~\cite{wang2020mead}, they fall short in providing accurate camera calibration and dynamic 3D models.

In this paper, we proposed a novel approach for synthesizing 3D talking heads with controllable emotion and an emotion-annotated calibrated multi-view video dataset for training our model. As shown in Fig.~\ref{fig:teaser}, the model renders high-fidelity free-view animations given the audio. The emotion labels used for driving can be set artificially and changed freely in time sequence. The detailed wrinkles and shadings caused by facial motions have been vividly synthesized. 
Such excellent performance benefits from two innovations, which are described below.

Firstly, we present a \textit{`Speech-to-Geometry-to-Appearance'} framework to map input speech to the dynamic appearance of a talking head. Specifically, a Speech-to-Geometry Network (S2GNet) is first used to predict 4D point clouds from audio features, where expression and lip motion are accurately recovered. Then, a 4D Gaussian model is established based on the predicted 4D points to represent the facial appearance efficiently. The appearance is disentangled into canonical Gaussians (static appearance) and dynamic Gaussians (facial motion-caused wrinkles, shading, etc). Geometry-to-Appearance Network (G2ANet) is introduced to learn the dynamic appearance from the multi-view videos and render free-view talking head animation. Experiments show that the proposed method generates more accurate and stable mouth shapes and models the dynamic details of facial motion, including wrinkles at specific facial expressions.

Secondly, we establish \datasetname dataset, an emotion-annotated multi-view talking head dataset with per-frame 3D facial shapes. \datasetname dataset are captured from $35$ subjects, and the data for each subject contains $20$ sentences under $8$ specific emotions with two emotional intensities, summing to $20$ minutes for each subject. 
We leverage an emotion extractor to parse emotion status from the input audio, then design an emotion-guided 4D prediction network to achieve emotion control over 3D talking head generation. The S2GNet is designed to be conditioned on emotion labels for emotion control, and G2ANet further synthesizes detailed expressions and dynamic details for specific input emotions. In this way, the emotion of our generated 3D talking head can be manipulated by manually set emotion labels.

The main contributions of this work can be summarized as:

\begin{itemize}
    \item We establish \datasetname dataset, an emotion-annotated multi-view talking head dataset with per-frame 3D facial shapes. Based on this unprecedented dataset, we propose the first explicit emotion-controllable 3D talking head synthesis method.
    \item A \textit{`Speech-to-Geometry-to-Appearance'} mapping framework is introduced to enhance lip synchronization and overall rendering quality of a 3D talking head.
    \item A 4D Gaussians model is proposed for representing the appearance of a 3D talking head, effectively synthesizing dynamic facial details such as wrinkles and subtle expressions. 
\end{itemize}

%% file: secs/2_related.tex
\section{Related Works}
\label{sec:related}

\subsection{Audio-driven Talking Head}

The key task of audio-driven talking heads is to establish correspondence between head motion and audio signals~\cite{brand1999voice,shiratori2006dancing,gan2020foley,ginosar2019learning}, and can be categorized into 2D-based and NeRF-based method according to heads' representation.

\textbf{2D-based Talking Head.}
Early learning-based talking heads~\cite{chung2017you, zhou2019talking, song2018talking} leverage the encoder-decoder architecture to synthesize a talking head video.
To further improve the lip-synchronization, some works~\cite{song2018talking, zhou2019talking, prajwal2020lip} extract disentangled appearance and semantic representations from speech or train an evaluation network targeted for synchronous lip movement. 
Later, generative models are introduced to produce talking heads directly from the extracted features or intermediate representation like facial landmarks~\cite{chen2019hierarchical, liao2020speech2video}.
Other 2D methods study how to edit full-frame videos, including the portrait's head, neck, and shoulder, often in dynamic backgrounds. Most existing methods~\cite{ezzat2002trainable, garrido2015vdub, sun2022masked, thies2020neural} synthesized the mouth-related area of the video and then blended it into the whole frame without altering other regions. 
Differently, other methods~\cite{suwajanakorn2017synthesizing} apply the re-timing strategy to find the optimal target frame with the predicted mouth shape or leveraged landmark as an intermediate representation~\cite{lu2021live,ji2021audio} with image-to-image translation networks to generate full-frame head animations.  
However, due to the lack of explicit 3D structure information, these methods fail to generate natural talking heads with consistent head poses.
 
\textbf{NeRF-based Talking Head.}
Recently, Neural Radiance Field (NeRF)~\cite{mildenhall2021nerf} has gained much attention for generating photo-realistic rendering of objects by using neural networks to learn the shape and appearance of the object under different spatial locations and viewpoints. Notably, Guo \etal~\cite{guo2021ad} proposed AD-NeRF, which utilizes audio-conditioned NeRF to synthesize the scene of a talking head for the first time. However, the head and the torso are learned from two sets of NeRF, leading to inconsistent results. To address this issue, Liu \etal~\cite{liu2022semantic} introduced SSP-NeRF, which incorporates an additional facial parsing brunch to improve the rendering efficiency and a torso deformation module to model the non-rigid torso deformations within a unified neural radiance field.  Additionally, DFA-NeRF~\cite{yao2022dfa} disentangled head pose, eye blink, and lip motion to synthesize personalized animation. To further accelerate training, DFRF~\cite{shen2022learning} learned a base model for lip motion that can be quickly fine-tuned to different identities, achieving a few-shot talking head synthesis. ER-NeRF~\cite{li2023efficient} resorts to optimizing the contribution of spatial regions by using a Tri-Plane Hash Representation. Differently, GeneFace~\cite{ye2023geneface} adopted a generative audio-to-motion model to improve the lip motion performance on out-of-domain audio. 
Different from all previous works, we leverage 3D Gaussian Splatting~\cite{kerbl20233d} to model dynamic facial motion while achieving superior visual quality with real-time rendering.

\subsection{Emotional Talking Head Generation}
\label{sec:related_emotion}
As a crucial factor in facial communication, emotion can enhance the authenticity of talking head animation. However, most previous researches focus on synchronizing lip motion with audio content, while neglecting facial expressions. Additionally, the absence of a large-scale audio-visual dataset with emotion annotations contributes to the challenge. To tackle this problem, Wang \etal~\cite{wang2020mead} introduced the MEAD dataset, comprising multi-view audio-visual data with eight emotions. Building upon this dataset, Ji \etal~\cite{ji2021audio} proposed EVP to decouple content and emotion information from the audio signal, facilitating facial emotion synthesis. Differently, some works resort to emotion labels~\cite{eskimez2021speech, wang2020mead, gan2023efficient, danvevcek2023emotional} or extract facial expressions from additional emotional videos~\cite{liang2022expressive, ji2022eamm, tan2023emmn, wang2023progressive} for emotion retrieval. In this work, we also go beyond mouth motion and expand into synthesizing facial expressions by extracting emotion and content from the audio signal to achieve precise emotion control.

\subsection{3D Talking Head Dataset}
\label{sec:related_dataset}
Most available talking head datasets have been captured in a controlled indoor environment. For example, Cudeiro \etal~\cite{cudeiro2019capture} collect VOCASET, a dataset of 4D head models created using 3D scanning and motion capture technology. Similarly, Wuu \etal~\cite{wuu2022multiface} present Multiface, which contains high-quality human head recordings under various facial expressions using a multi-view capture stage. Meanwhile, Pan \etal~\cite{pan2024renderme} produce RenderMe-360 with rich annotations including different granularities. FaceScape~\cite{zhu2023facescape, yang2020facescape} proposed to collect 3D faces in multiple standardized expressions for each subject to generate riggable 3D blendshapes, which can be driven by 3DMM coefficients to generate 3D talking face videos~\cite{zhuang2022mofanerf, he2024head360}. However, this strategy has poor accuracy in modeling lip and facial expressions, due to the limited representation ability of Blendshapes.
Other studies~\cite{alghamdi2018corpus,wang2020mead} have also presented multi-view audio-visual data with the potential to reconstruct 3D heads. However, only front and side view data were provided in Lombard~\cite{alghamdi2018corpus}. While MEAD~\cite{wang2020mead} recorded emotional audio-visual clips at seven views, they lack camera calibration data, which brings unstable reconstruction results. Another type of work like HDTF~\cite{zhang2021flow} collected a large in-the-wild audio-visual dataset and leverages the 3D morphable model (3DMM) to fit the monocular video, but the fitting results are typically coarse.  Different from all prior works above, our work presents the \datasetname dataset, an emotion-annotated 4D dataset with speeches and per-frame 3D facial shapes. 

%% file: secs/3_dataset.tex
\section{Dataset}
\label{sec:emodata}

We present \datasetname dataset, an emotion-annotated multi-view face dataset with reconstructed 4D face models and accurate camera calibrations. The dataset contains $30$ subjects, and the data for each subject contains $20$ sentences under $8$ specific emotions with two emotional intensities, summing to roughly $20$ minutes for each subject. The collected emotions include `neutral', `angry', `contempt', `disgusted', `fear', `happy', `sad', and `surprised'. Except for `neutral', two intensities - `mild' and `strong' are captured for each emotion. We invited professional performance instructors to guide the subjects in expressing the correct emotions. 

For data acquisition, we built a dome with $11$ video cameras in the same horizontal plane as the subject's head and evenly around the head in a $180^\circ$ degree focusing on the frontal face.  All cameras are temporally synced with a hardware synchronization system and are calibrated before capturing videos. State-of-the-art multi-view 3D reconstruction algorithm~\cite{zhang2022vismvsnet} is leveraged to reconstruct accurate 3D triangle mesh models, which are then transformed into topologically uniformed 3D mesh models~\cite{amberg2007optimal}. The 3D vertices of the 3D surface model corresponding to each frame constitute the 3D points stream, namely 4D points, which are used for training our 3D talking face model. 

To the best of our knowledge, \datasetname dataset is the first talking face dataset that contains both emotion annotations and per-frame 3D facial geometry. The meta parameters of our dataset and previous talking face datasets are compared in Table~\ref{tab:dataset_comp}. The dataset has been publicly released for research purposes on our project page. More details about the \datasetname dataset are explained in the supplementary material.

\begin{table}[t]
\setlength{\belowcaptionskip}{0.1cm}
\caption{Comparison of 3D Talking Head Datasets}
\centering
%\resizebox{0.8\textwidth}{!}{
\begin{tabular}{@{}lccccc@{}}
\toprule
Dataset        & Identities & Camera Num. & Duration  & 3D Shape       & Emotion        \\ \midrule
VOCASET~\cite{pan2022vocal} & 12      & 6+12$^*$& 29min& \usym{1F5F8}   & \usym{2715}    \\
Multi-Face~\cite{wuu2022multiface}     & 13      & 80        & 7min& \usym{1F5F8}   & \usym{2715}    \\
 BIWI~\cite{fanelli2013random} & 20& 1& -& \usym{2715}    &\usym{2715}    \\
MEAD~\cite{wang2020mead}           & 60& 7& 39min& \usym{2715}    & \usym{1F5F8}   \\
 RenderMe-360~\cite{pan2024renderme} & 500& 60& -& \usym{1F5F8}  &\usym{2715}    \\
Ours           & 30      & 11        & 20min      & \usym{1F5F8}  & \usym{1F5F8}   \\ \bottomrule
\multicolumn{6}{l}{\small $^*$ means 6 3D scanners and 12 RGB cameras are used.}\\
\end{tabular}
%}
\label{tab:dataset_comp}
\end{table}

\begin{figure}[t]
  \centering
  \setlength{\abovecaptionskip}{0.2cm}
  \setlength{\belowcaptionskip}{-0.3cm}
  \includegraphics[width=1.0\linewidth]{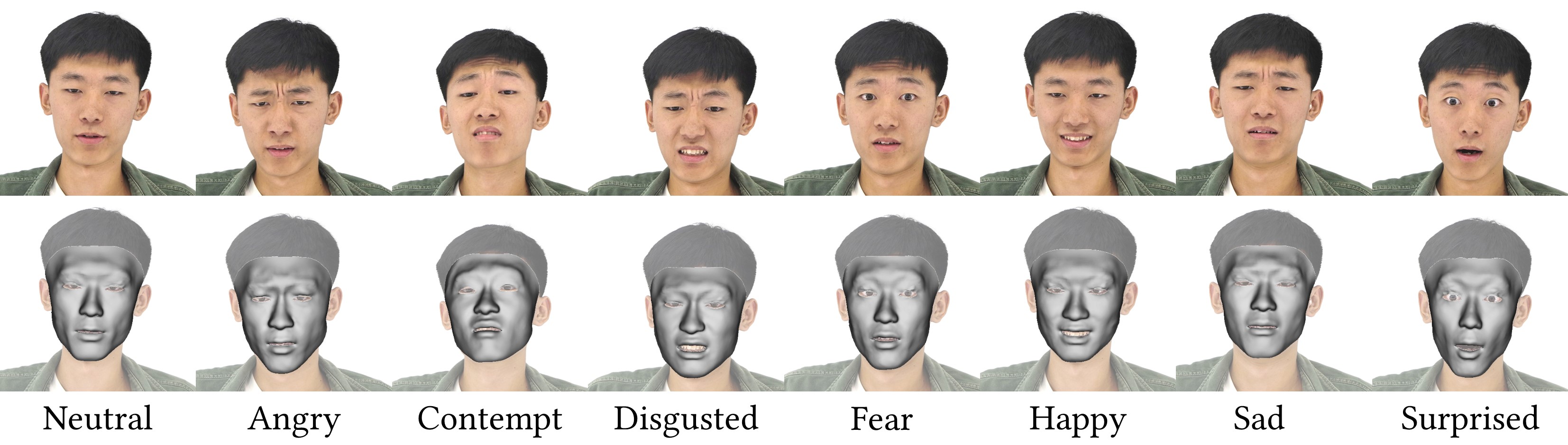}
  \caption{ \textbf{\datasetname Dataset.} We collect a multi-view talking face dataset, where each subject's data contains $8$ emotions and the reconstructed 3D mesh model for each frame. It is worth noting that the provided per-frame 3D models cannot represent detailed 3D shapes like wrinkles but can be learned from videos by our G2ANet (Sec.~\ref{sec:g2a}).
  }
  \label{fig:dataset}
\end{figure}

%% file: secs/4_method.tex
\section{Method}
\label{sec:method}

\begin{figure*}[t]
  \centering
  \setlength{\abovecaptionskip}{0.2cm}
  \setlength{\belowcaptionskip}{-0.3cm}
  \includegraphics[width=1.0\linewidth]{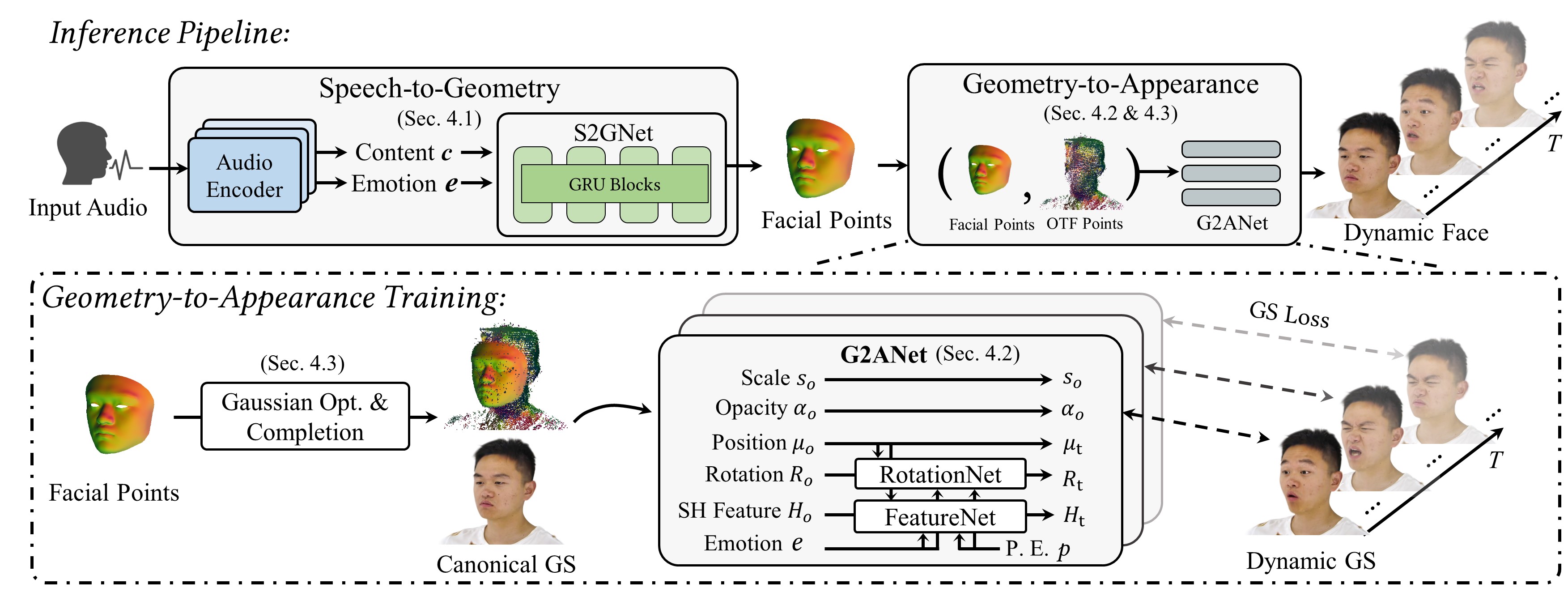}
  \caption{ \textbf{Overall Pipeline.} The pipeline consists of five modules: 1) Audio Encoder that parses content features from input speech; 2) Speech-to-Geometry Network (S2GNet) that predicts dynamic 3D point clouds from the features; 3) Gaussian Optimization and Completion Module for establishing a canonical appearance; 4) Geometry-to-Appearance Network (G2ANet) that synthesizes facial appearance based on dynamic 3D point cloud; and 5) Rendering module for rendering dynamic Gaussians into free-view animations. 
  }
  \label{fig:pipeline}
\end{figure*}

As shown in Fig.~\ref{fig:pipeline}, our method consists of the following modules: 1) the Audio Enocoder that encodes audio features and extracts emotional labels from input speech; 2) Speech-to-Geometry Network (S2GNet) that predicts dynamic 3D point clouds from the audio features and emotional labels; 3) Static Gaussians Optimization and Completion Module for establishing a canonical appearance model; 4) Geometry-to-Appearance Network (G2ANet) that synthesizes facial appearance based on dynamic 3D point cloud. The above modules together constitute the \textit{`Speech-to-Geometry-to-Appearance'} mapping framework for emotional talking head synthesis.

%In the following, we will first explain the \textit{Speech-to-Geometry} method in Sec.~\ref{sec:s2g}. Then, Sec.~\ref{sec:g2a} will introduce the \textit{Geometry-to-Appearance} model, along with the utilization of 4D Gaussians for generating photo-realistic 3D talking faces with dynamic details. Lastly, in Sec.~\ref{sec:head_compl}, we explain how to synthesize the entirety of a talking head, extending beyond the facial synthesis to hair, neck, and shoulders.

\subsection{Speech-to-Geometry}
\label{sec:s2g}

As shown in Fig.~\ref{fig:pipeline}, the speech-to-geometry module consists of two parts: an audio encoder that converts the input speech into audio features and the S2GNet that converts audio features and emotional labels into 3D mesh sequences. Concretely, a pre-trained HuBERT speech model~\cite{hsu2021hubert} is used as the audio encoder. 
Drawing inspiration from Baevski~\etal's method~\cite{baevski2020wav2vec}, we employ a context network adhering to the transformer architecture~\cite{vaswani2017attention} as the backbone of our emotion extractor. The model is initialized with the pre-trained weights from Baevski~\etal's work and then fine-tuned with ground-truth emotional labels and speech audio in our dataset.   

The design of S2GNet follows FaceXHubert~\cite{haque2023facexhubert}, a cutting-edge audio-to-mesh prediction network. Specifically, S2GNet receives extracted emotion labels and encoded audio features as input, regressing vertex displacements based on the mean template mesh, ultimately producing the 4D talking mesh sequence. Rather than relying on the commonly used transformer-based networks, S2GNet opts for the Gated Recurrent Unit (GRU)~\cite{cho2014properties} as its core architecture. Through experiments, we have verified that this approach achieves superior performance in lip-synchronization and speech generalization. 

We leverage FaceXHubert's pre-trained model to initialize the training of S2GNet, then finetune S2GNet on our \datasetname dataset. The triangle mesh model predicted by S2GNet was quadrupled upsampled, which is then used as the input to the geometry-to-appearance module.  More details about the training and networks are illustrated in the supplementary material.

\subsection{Geometry-to-Appearance}
\label{sec:g2a}

After predicting the 4D facial points, we introduce the Geometry2Appearance Network (G2ANet) to synthesize the 3D Gaussian features~\cite{kerbl20233d} of the talking face by taking 4D points as input, as shown in Fig.~\ref{fig:pipeline}. 

The design of G2ANet is founded upon two key observations. Firstly, we recognize that for a specific individual's talking head, facial movements produce relatively minor variations in appearance. With this in mind, we initially train a Gaussian model on a static, non-speaking head, referred to as Canonical Gaussians. Subsequently, a neural network is employed to predict appearance changes caused by taking motions, using Canonical Gaussians and 4D point-derived facial movements as inputs. These predicted variations are named Dynamic Gaussians. Secondly, given that there is no direct correlation between speech and non-facial features such as hair, neck, and shoulders, the S2GNet Network disregards predicting the geometric structure of these elements. Consequently, G2ANet learns to replicate the appearance of these non-facial parts and seamlessly integrates them with the face, as elaborated in Section~\ref{sec:head_compl}.

\textbf{Canonical 3D Gaussians.}
Following 3D Gaussian Splatting (3DGS)~\cite{kerbl20233d}, we represent the static head as 3D Gaussians that are parameterized 3D points. Each 3D Gaussian is represented by the 3D point position $\mu$ and covariance matrix $\mathbf{\Sigma}$, and the density function is formulated as:

\begin{equation}
    g(x) = e^{-\frac{1}{2}(x-\mu)^T \mathbf{\Sigma}^{-1}(x-\mu)}
\end{equation}

As 3D Gaussians can be formulated as a 3D ellipsoid, the covariance matrix $\mathbf{\Sigma}$ is further formulated as:

\begin{equation}
    \mathbf{\Sigma} = \mathbf{R}\mathbf{S}\mathbf{S}^T\mathbf{R}^T
\end{equation}

\noindent where $\mathbf{S}$ is a scale and $\mathbf{R}$ is a rotation matrix. The 3D Gaussians are differentiable and can be easily projected to 2D splats for rendering. In the differentiable rendering phase, $g(x)$ is multiplied by an opacity $\alpha$, then splatted onto 2D planes and blended to constitute colors for each pixel. The appearance is modeled with an optimizable $48$-dimension vector $\mathbf{H}$ representing four bands of spherical harmonics. In this way, the appearance of a static head can be represented as 3D Gaussians $\mathbf{G}$:

\begin{equation}
    \mathbf{G} \leftarrow \{\mu, \mathbf{S}, \mathbf{R}, \alpha, \mathbf{H}\}
\end{equation}

\noindent where $\leftarrow$ means $G$ is a set of parameterized points, each represented by a parameter set on the right of the arrow.

In our approach, the canonical 3D Gaussians $\mathbf{G}_o$ represent a static head avatar and are learned from multi-view images of a moment without speech, usually the first frame of a video clip. The canonical 3D Gaussians $\mathbf{G}_o$ are denoted as:

\begin{equation}
    \mathbf{G}_o \leftarrow \{\mu_o, \mathbf{S}_o, \mathbf{R}_o, \alpha_o, \mathbf{H}_o\}
\end{equation}

\textbf{Dynamic Detail Synthesis.} To generate a 3D talking face with 3D Gaussians, we propose to predict the appearance at $t$ moment with dynamic details. The dynamic details mean the detailed appearance due to the talking facial motion, such as specific wrinkles and subtle expressions. For a specific subject, the opacity $\alpha$ and scale $\mathbf{S}$ remain unchanged and equal to $\alpha_o$ and $\mathbf{S}_o$ respectively. The Gaussians at $t$ time $\mathbf{G}_t$ is formulated as:

\begin{equation}
    \mathbf{G}_t \leftarrow \{\mu_t, \mathbf{S}_o, \mathbf{R}_t, \alpha_o, \mathbf{H}_t\}
\end{equation}

As $\mu_t$ has been predicted in the speech-to-geometry network, only $\mathbf{H}_t$ and $\mathbf{R}_t$ are unknown and are predicted by FeatureNet and RotationNet, respectively:

\begin{equation}
    \mathbf{H}_t = FeatureNet(\mathbf{H}_o, e, p, \delta \mu)
\end{equation}

\begin{equation}
    \mathbf{R}_t = RotationNet(\mathbf{R}_o, e, p, \delta \mu)
\end{equation}

\noindent where $e$ is the emotion vector; $p$ is the position in the UV coordinate predefined for each point; $\delta \mu$ is defined as:

\begin{equation}
\delta \mu = \mu_t - \mu_o - \frac{1}{N}\sum_{i \in G}{(\mu_t^i - \mu_o^i)}
\end{equation}

FeatureNet and RotationNet are 4-layer and 2-layer MLPs, respectively. The detailed architectures of the two networks are reported in the supplementary material. 

Through experiments, we find that combining canonical Gaussians with dynamic Gaussians enables the prediction of highly detailed dynamic facial appearance. These include wrinkles formed during angry or other expressions and nuanced appearance variations resulting from lip movements. No preceding talking head model has achieved such meticulous rendering of dynamic facial details.

\begin{figure}[t]
  \centering
  \includegraphics[width=1.0\linewidth]{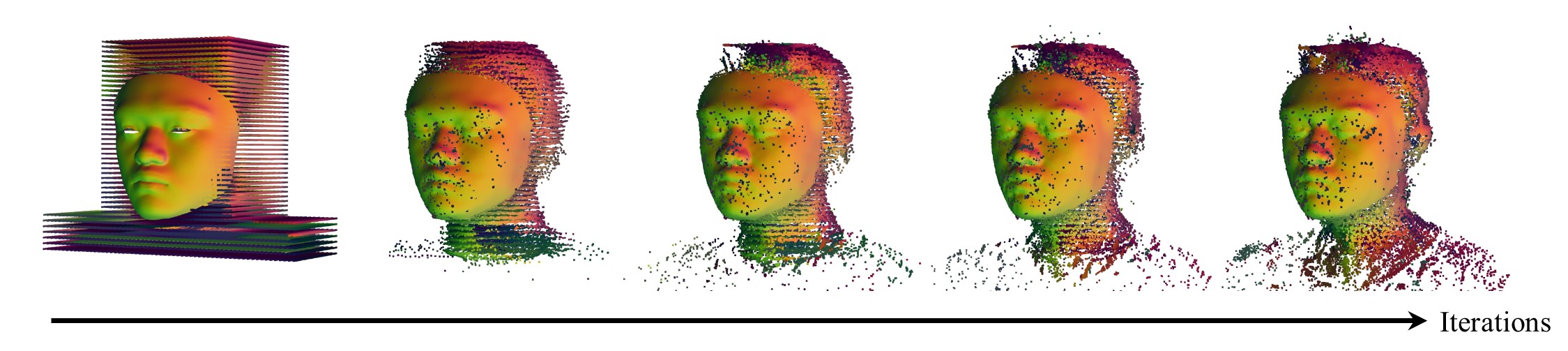}
  \caption{ \textbf{Points Completion.} 
  S2GNet solely generates the point cloud for the facial region. In contrast, points other than the facial region (OTF points) are optimized from uniformly initialized points. This figure illustrates the gradual optimization process of the OTF points, culminating in forming a complete head structure. 
  }
  \label{fig:compl}
\end{figure}

\subsection{Head Completion} 
\label{sec:head_compl}

The speech-to-geometry network only predicts the 3D point cloud of the face, as the geometry other than the face, such as hair, neck, and shoulders, are not strongly correlated to the speech signals. Therefore, the regions other than the face do not have accurate initial points for optimizing Gaussians, so we add $85,500$ uniformly distributed points in the space to constitute the regions other than the face by 3DGS optimization, as shown in Fig.~\ref{fig:compl}.

The points other than the facial region (OTF points for short), including hair, neck, and shoulders, should follow the face point motion to a certain extent. We observed that these points tend to be stationary at points far from the face, such as the shoulders, while points close to the face will move together, such as the hair and neck. So we formulate the motion of the points except for the facial points $\delta \mu_{otf}$ as:

\begin{equation}
    \delta \mu_{otf} = \delta \mu_{f} \cdot e^{-\alpha d}
\end{equation}

\noindent where $\delta \mu_{f}$ is the movement of a facial point that is closest to this OTF point; $r$ is the distance between the OTF point and its closest facial point; $\alpha$ is the decay factor and is set to $0.1$ in all our experiments.

In the training of dynamic Gaussians, we observed that training the Gaussians on hair and shoulders the same way as we did for faces would cause severe blur. It is because the facial points' position in our dataset is accurately recovered, so the Gaussian points can be more accurately aligned on every frame within a video clip. By contrast, the appearance of regions outside the face, including hair, neck, and shoulders, is optimized from random points. The 3D positions of these OTF points are unreliable, so these points will correspond to misaligned appearances at different frames, leading to a vague appearance predicted. On the other side, the complete head represented by the canonical Gaussians is clear, however, the appearance of the canonical Gaussians is static.

To solve this problem, we propose to treat the dynamic part (the face) and the static part (the area above the shoulders other than the face) separately, using dynamic Gaussians for the former and canonical Gaussians for the latter. Specifically, we pre-define a natural transition facial weight mask, which is applied to opacity $\alpha$ to achieve a natural fusion of dynamic and static parts. The weight $W_p$ for point $p$ is formulated as:

\begin{equation}
    W_p = \left\{
    \begin{array}{lr}
    0, \qquad \qquad d<d_{0}\\
    \frac{d-d_{0}}{d_{th}}, \qquad d_{0} \le d \le d_{0}+d_{th}\\
    1, \qquad \qquad d_{0}+d_{th} < d
    \end{array}
    \right.
\end{equation}

\noindent where $d$ is the distance between point $p$ and its closest facial point; $d_0$ and $d_{th}$ are transition factors and are set to $5mm$ and $10mm$, respectively. 
The opacity field for canonical Gaussians and dynamic Gaussians are multiplied by $W_p$ and $(1-W_p)$, respectively. 
In this way, the talking facial appearance and the other appearance (hair, neck, and shoulders) are fused to synthesize a clear and complete 3D talking head.

%% file: secs/5_exp.tex
\section{Experiments}
\label{sec:exp}

\subsection{Implementation Details}
\label{sec:details}

\noindent \textbf{Data pre-processing.} Before training our model, we crop and resize the raw frames into a squared image at resolution $512\times512$. A matting algorithm~\cite{chen2022pp} is used to remove the background. For each subject, we randomly divided the data into a training set and a test set in a ratio of 4:1.

\noindent \textbf{Training details.} 
Adam\cite{kingma2014adam} optimizer is adopted to train the speech-to-geometry network (S2GNet) and geometry-to-appearance network (G2ANet). 
In the training of G2ANet, we first obtain canonical Gaussians by training with the 1st frame of the video with neutral expression and closed mouth. The number of facial points and the OTF points are $100$ and $200$, respectively. The clone and splitting are prohibited to ensure a stable correspondence association of points between frames. Then, FeatureNet and RotationNet are trained to predict the dynamic details. Due to space limitations, more implementation details are explained in the supplementary material.

\begin{figure}[ht]
  \centering
  \includegraphics[width=1.0\linewidth]{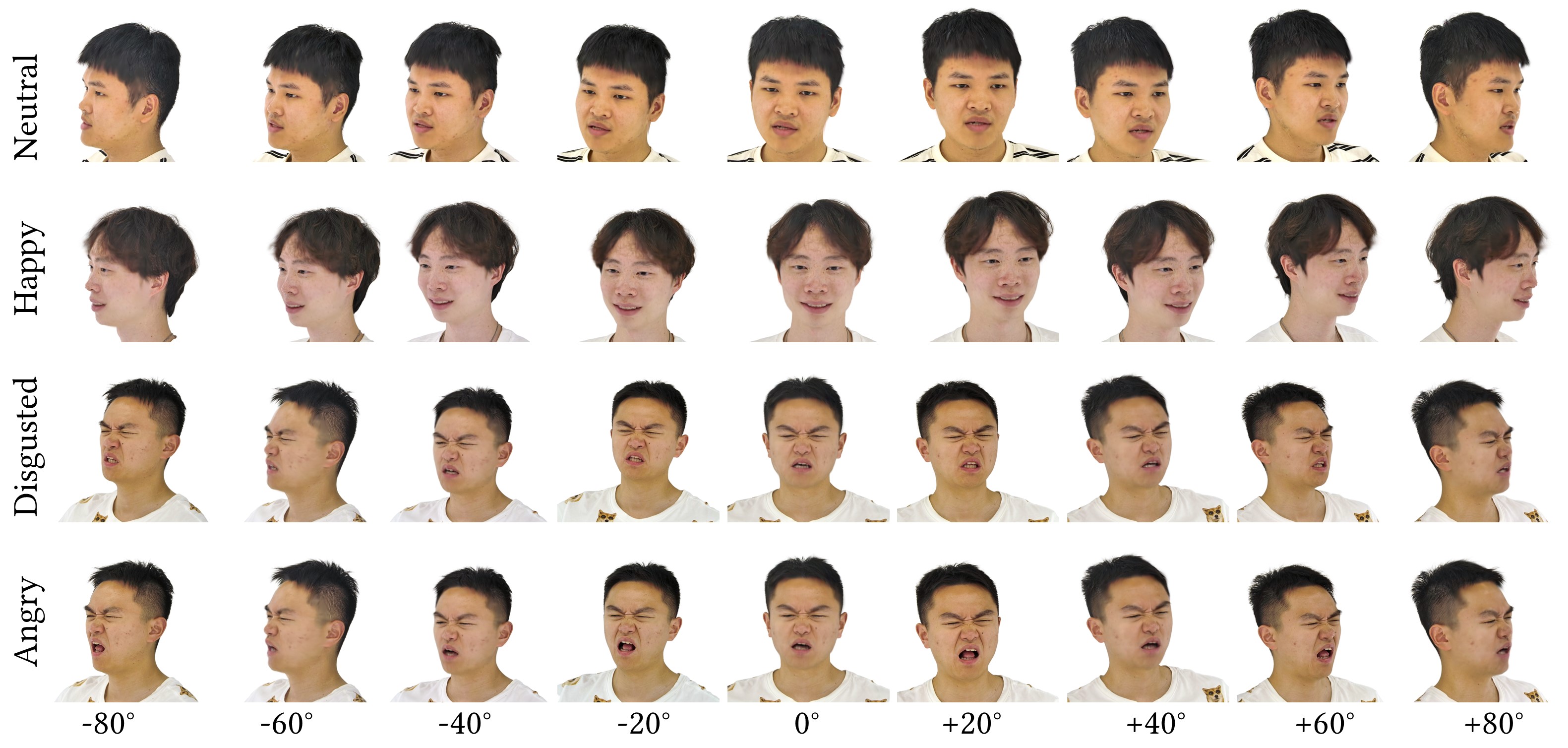}
  \caption{ \textbf{Multi-view Synthesis and Emotional Control.} 
  }
  \label{fig:self_qual}
\end{figure}

The renderings of our results at different views with different emotions are shown in Fig.~\ref{fig:self_qual}.  It can be seen that high-quality frames are rendered at large-angle views with correct emotion synthesized. We recommend watching the supplementary video to evaluate lip-sync and rendering performance qualitatively.

We compare our method with several image-based methods (MakeItTalk~\cite{zhou2020makelttalk}, EAMM~\cite{ji2022eamm}, SadTalker~\cite{zhang2023sadtalker}, DreamTalk~\cite{ma2023dreamtalk}), three 3D methods (AD-NeRF~\cite{guo2021ad}, Real3D-Portrait~\cite{ye2024real3d}, Next3D~\cite{sun2023next3d}), and two speech-to-mesh methods (VOCA~\cite{cudeiro2019capture}, MeshTalk~\cite{richard2021meshtalk}). PSNR~\cite{hore2010image} and SSIM\cite{wang2004image} are adopted to evaluate the overall image quality and LPIPS\cite{zhang2018unreasonable} is adopted to measure the perceptual similarity between the results and the ground truth. Landmark Distance (LMD)\cite{chen2018lip} is utilized to evaluate lip synchronization, and Cumulative Probability of Blur Detection (CPBD) \cite{narvekar2009no} is used to measure the sharpness of the result images.

\begin{figure*}[t]
  \centering
  \includegraphics[width=1.0\linewidth]{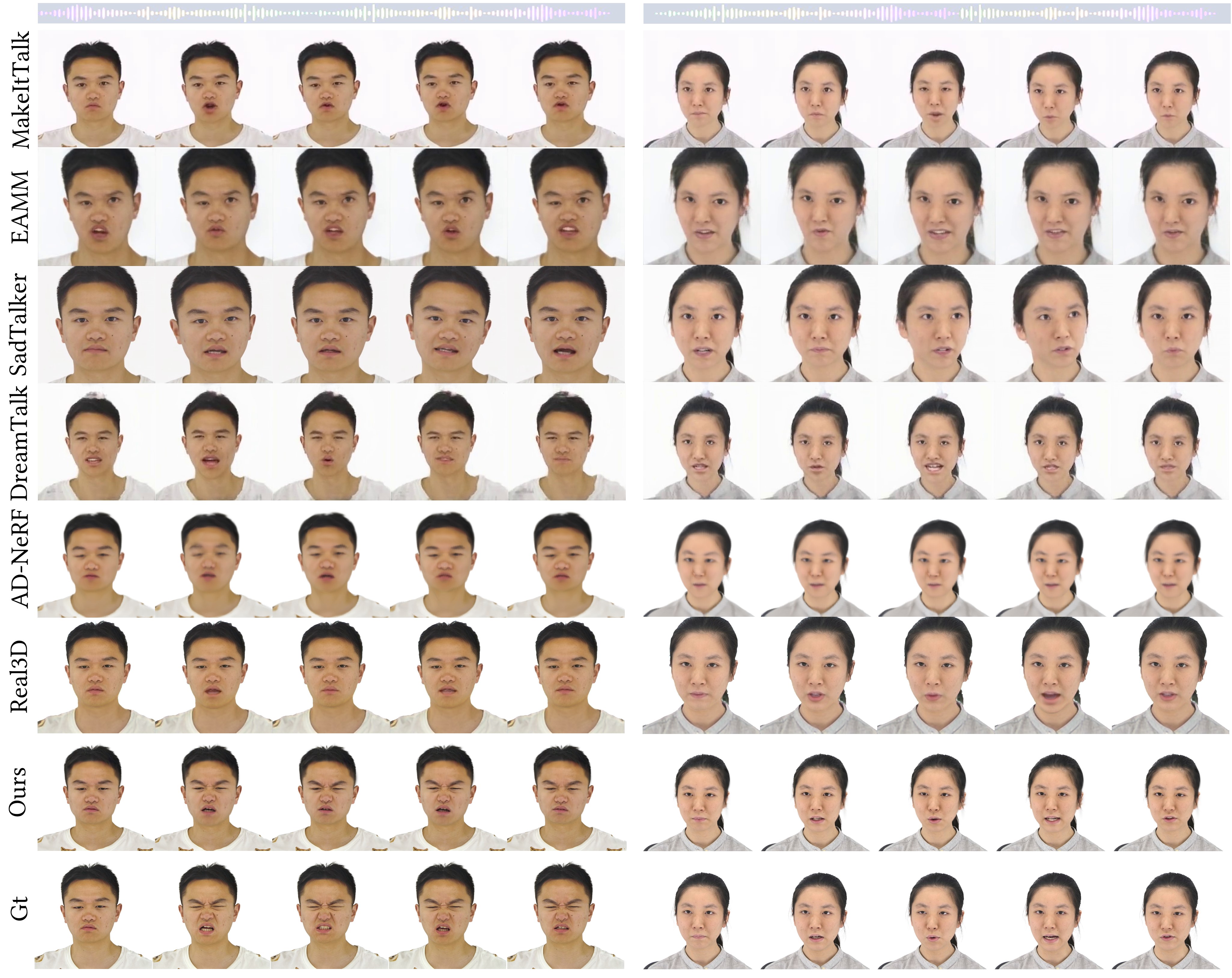}
  \caption{ \textbf{Qualitive Comparison.}  Our method outperforms previous works in both lip synchronization and rendering quality. Besides, our method is the only one that is wide-range view renderable and emotion-controllable. %We recommend watching our supplemental video for better qualitative assessment.
  }
  \label{fig:comp}
\end{figure*}

\begin{table}[t]
\centering
\setlength{\abovecaptionskip}{-0.2cm}
\setlength{\belowcaptionskip}{0.1cm}
\caption{\textbf{Quantitative Comparison}}
\resizebox{0.7\textwidth}{!}{
\begin{tabular}{@{}lccccc@{}}
\toprule
\multicolumn{1}{c}{Method} & PSNR $\uparrow$ & SSIM $\uparrow$ & LPIPS $\downarrow$ & LMD $\downarrow$  & CPBD $\uparrow$ \\ \midrule
MakeItTalk~\cite{zhou2020makelttalk}                 & 20.61 & 0.79 & \textbf{0.12}  & 4.45 & 0.29                     \\
EAMM~\cite{ji2022eamm}                       & 9,82 & 0.57 & 0.40  & 20.25 & 0.12                     \\
SadTalker~\cite{zhang2023sadtalker}                  & 9.90 & 0.64 & 0.47  & 43.49 & 0.11                     \\
DreamTalk~\cite{ma2023dreamtalk}                  & 19.19 & 0.76 & 0.17  & 3.76 & \textbf{0.38}                     \\
AD-NeRF~\cite{guo2021ad}                    & 20.39 & \textbf{0.83} & 0.22  & 5.10 & 0.04                     \\
Real3D-Portrait~\cite{ye2024real3d}            & 13.53 & 0.72 & 0.30  & 24.11 & 0.26                     \\
Ours                       & \textbf{21.22} & \textbf{0.83} & \textbf{0.12}  & \textbf{3.62} & 0.30                     \\ \bottomrule
\end{tabular}
}
\label{tab:quan}
\end{table}

\subsection{Qualitative and Quantitative Evaluations}

As shown in Fig.~\ref{fig:comp} and Tab.~\ref{tab:quan}, our results outperform all previous 2D image-based and 3D-based talking heads.  More importantly, our result can be rendered at $-90^\circ - +90^\circ$ and explicitly emotion-controllable, as shown in Fig.~\ref{fig:self_qual}, which all other methods cannot achieve. Our method also leads in the scores of PSNR, SSIM, LPIPS, and LMD, which indicates that our method improves rendering quality and lip-sync compared to previous works.

\noindent \textbf{User Study.} 
We conduct a user study to evaluate the quality of the synthesized portraits by comparing the real data with the generated ones from different approaches. Specifically, we sample 10 generated videos from the test set with various identities and viewpoints. $27$ participants are invited to rate these videos with a score from 1 (worst) to 5 (best) w.r.t three aspects: speech-visual synchronization, video fidelity, and image quality. A more detailed explanation of these rating criteria is provided in the supplementary material. As VOCA and MeshTalk only produce mesh with no texture, the image quality for them is not evaluated. The results are shown in Table~\ref{tab:user}.  Our method obtains the highest score on all aspects, indicating better performance in speech-visual synchronization, video fidelity, and image quality.

\begin{table}[t]
\centering
\setlength{\belowcaptionskip}{0.1cm}
\caption{\textbf{User Study.}}
\resizebox{0.8\textwidth}{!}{
\begin{tabular}{@{}lcccc@{}}
\toprule
\multicolumn{1}{c}{Method} & Speech-Visual Sync & Video Fidelity & Image Quality \\ \midrule
MakeItTalk~\cite{zhou2020makelttalk}                 & 2.88               & 2.50            & 2.75               \\
EAMM~\cite{ji2022eamm}                       & 1.88               & 1.88            & 2.14                       \\
SadTalker~\cite{zhang2023sadtalker}                  & 3.13               & 3.25            & \textbf{4.25}               \\
DreamTalk~\cite{ma2023dreamtalk}                  & 3.00               & 1.89            & 1.75                  \\
AD-NeRF~\cite{guo2021ad}                    & 1.65               & 2.12            & 1.85                        \\
Real3D-Portrait~\cite{ye2024real3d}            & 3.50               & 2.38            & 3.25                     \\
Next3D~\cite{sun2023next3d}                     & 2.37               & 2.39            & 3.75                    \\
VOCA~\cite{cudeiro2019capture}                       & 3.54               & 2.50            & /                  \\
MeshTalk~\cite{richard2021meshtalk}                   & \textbf{3.60}               & 2.40            & /        \\
%GaussianAvatars~\cite{qian2023gaussianavatars}            & 0.00               & 0.00            & 0.00           & 0.00                    \\
Ours                       & 3.54               & \textbf{3.96}            & \textbf{4.25}                       \\ \bottomrule
\end{tabular}
}
\label{tab:user}
\end{table}

\subsection{Ablation Study}

\begin{figure}[t]
  \centering
  \setlength{\abovecaptionskip}{0.2cm}
  \setlength{\belowcaptionskip}{-0.2cm}
  \includegraphics[width=0.7\linewidth]{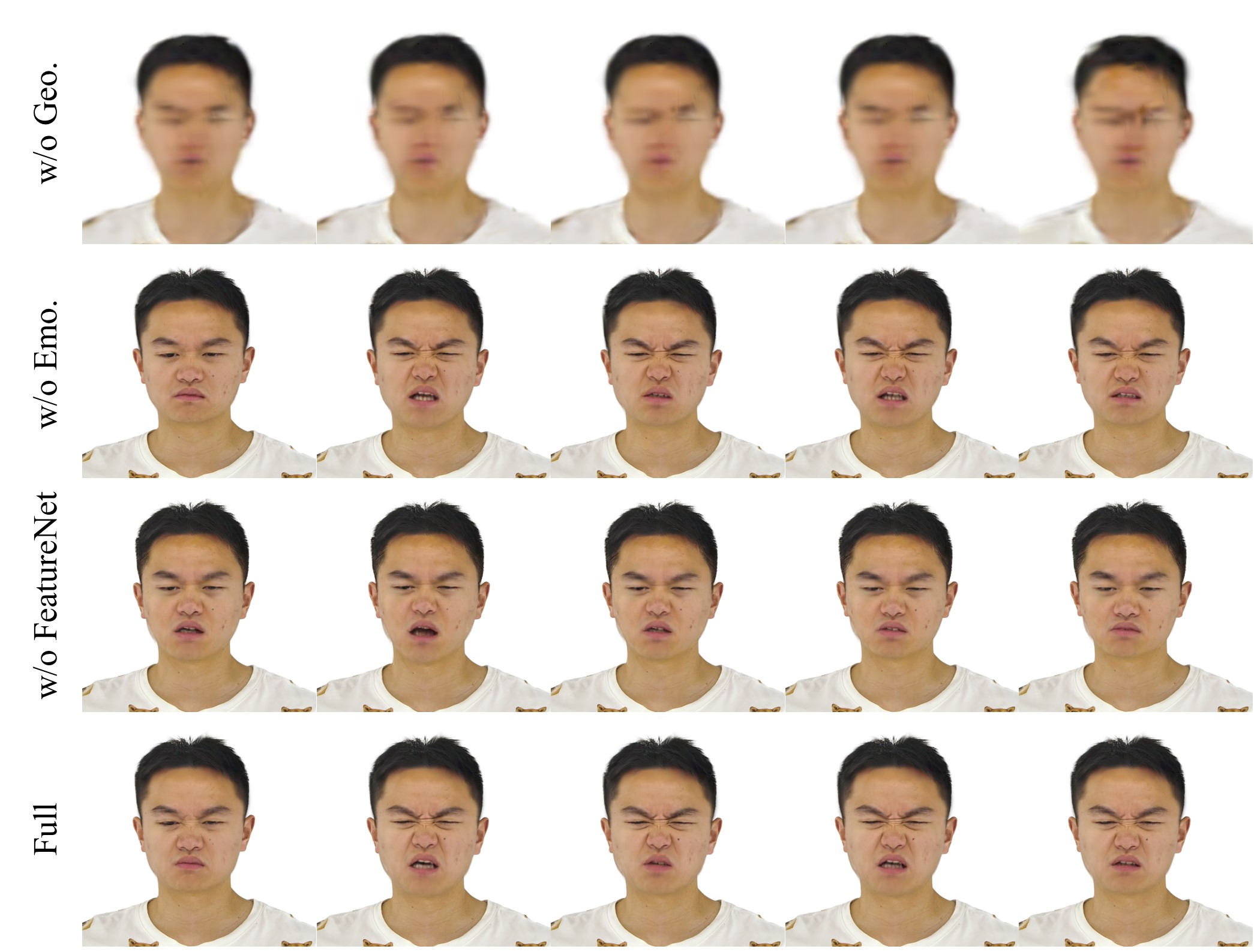}
  \caption{ \textbf{Ablation Study.} The effectiveness of the proposed modules is verified. %The full method has a clear advantage in image quality and facial details.
  }
  \label{fig:ablation}
\end{figure}

\begin{table}[t]
\centering
\setlength{\belowcaptionskip}{0.1cm}
\caption{\textbf{Ablation Study.}}
\resizebox{0.7\textwidth}{!}{
\begin{tabular}{@{}lccccc@{}}
\toprule
\multicolumn{1}{c}{Method} & PSNR $\uparrow$ & SSIM $\uparrow$ & LPIPS $\downarrow$ & LMD $\downarrow$  & CPBD $\uparrow$ \\ \midrule
w/o Geometry            & \textbf{24.30}$^*$ & \textbf{0.84}$^*$ & 0.29  & 10.83 & 0.06                     \\
w/o Emotion         & 21.19 & 0.82 & 0.13  & 5.60 & 0.28                     \\
%w/o P. E.           & 21.14 & 0.82 & 0.13  & 5.77 & 0.29                     \\
w/o FeatureNet      & 20.85 & 0.81 & \textbf{0.12}  & 4.93 & \textbf{0.35}                     \\
Full                & \textbf{21.22} & \textbf{0.83} & \textbf{0.12}  & \textbf{3.62} & 0.30                     \\ \bottomrule
\end{tabular}
}
\label{tab:ablation}
\end{table}

We conducted three sets of ablation studies to analyze the effectiveness of each module in our method. The comparison results are shown in Fig.~\ref{fig:ablation}.

\noindent $\bullet$ \textbf{(A) w/o Geometry.} We compare the \textit{`speech-to-geometry-to-appearance'} framework with the traditional `speech-to-appearance' framework to analyze the effectiveness of the former. In the `speech-appearance' framework, all parameters of 3D Gaussians, that is Dynamic 3D Gaussians $\mathbf{G}_t \leftarrow \{\mu_t, \mathbf{S}_o, \mathbf{R}_t, \alpha_o, \mathbf{H}_t\}$, are directly predicted from the audio features using several networks. We adopted the same design of FeatureNet and RotationNet to build the `speech-to-appearance' model, only made changes to how we obtain $\mu_{t}$ - a network with comparable parameters of S2GNet is used to predict $\delta \mu$ to calculate $\mu_{t}$, and is trained in an end-to-end manner.  
As shown in Fig.~\ref{fig:ablation} and Tab.~\ref{tab:ablation}, introducing geometry leads to much higher image quality. Though PSNR and SSIM scores of `w/o Geometry' are higher, the rendering results clearly show that the facial details are all missing. We believe that this is because the introduction of geometry decomposes the complex \textit{Speech-to-Appearance} mapping problem into two more stable and straightforward mappings - \textit{Speech-to-Geometry} and \textit{Geometric-to-Appearance}. This strategy reduces the complexity of 3D talking head synthesis, which makes it easier for neural networks to learn. 

\noindent $\bullet$ \textbf{(B) w/o Emotion.} The proposed method leverages explicit extraction of emotional components in speech for generating talking face animation (Section~\ref{sec:s2g}). We try to remove the emotion extractor and emotion labels in the S2GNet to evaluate the performance without emotion extraction and embedding.  As shown in Fig.~\ref{fig:ablation} and Table~\ref{tab:ablation}, the target emotion is not accurately synthesized in the generated videos after removing emotion-related designs, demonstrating the effectiveness of emotion extracting and encoding. 

\noindent $\bullet$ \textbf{(C) w/o FeatureNet.} In G2ANet (Section~\ref{sec:g2a}), we propose to leverage FeatureNet to model the dynamic appearance. Here, we remove the FeatureNet from the design and evaluate the performance. As shown in Fig.~\ref{fig:ablation}, the dynamic wrinkles are weakened and the expressions tend to be neutral. The overall facial appearance is degraded towards a canonical appearance, which demonstrates the effectiveness of the FeatureNet in G2ANet.

%% file: secs/6_con.tex
\section{Conclusion}
\label{sec:con}

In this paper, we propose to synthesize a high-fidelity, emotion-controllable 3D talking head that can be rendered over a wide range of viewing angles. A \textit{`Speech-to-Geometry-to-Appearance'} mapping framework with dynamic 4D Gaussians is proposed for better lip-sync and rendering quality. A multi-view video dataset with emotion annotation and per-frame 3D facial shapes is presented for learning 3D talking heads. Our data and algorithm can be the foundation for future research about emotion-controllable 3D talking heads.

\textbf{Limitation.} Our model is person-specific, which means only one identity can be generated by training a neural network. Therefore, our method relies on a well-calibrated multi-view camera system to collect videos for training, and cannot synthesize human appearance from a single image. In addition, our method fails to model dynamic flapping hair, which is also a challenging problem.

\noindent\textbf{Acknowledgements} This study was funded by NKRDC 2022YFF0902400, NSFC 62441204, and Huawei.